# AI-Assisted Decision-Making for Clinical Assessment of Auto-Segmented Contour Quality

Biling Wang[1,2], Austen Maniscalco[1,3], Ti Bai[1,3], Siqiu Wang[1,3], Michael Dohopolski[1,3], Mu-Han Lin[1,3], Chenyang Shen[1,3], Dan Nguyen[1,3], Junzhou Huang[4], Steve Jiang[1,3,*], Xinlei Wang[5,6,*]

[1]Medical Artificial Intelligence and Automation Laboratory, University of Texas Southwestern Medical Center, Dallas, Texas, USA
[2]Department of Statistics and Data Science, Southern Methodist University, Dallas, Texas, USA
[3]Department of Radiation Oncology, University of Texas Southwestern Medical Center, Dallas, Texas, USA
[4]Department of Computer Science and Engineering, University of Texas at Arlington, Arlington, Texas, USA
[5]Department of Mathematics, University of Texas at Arlington, Arlington, Texas, USA
[6]Division of Data Science, College of Science, University of Texas at Arlington, Arlington, Texas, USA
*Co-Correspondence Authors. Email: steve.jiang@utsouthwestern.edu; xinlei.wang@uta.edu

## Abstract

**Purpose:** This study introduces a novel Deep Learning (DL)-based quality assessment (QA) approach specifically designed for evaluating auto-generated contours (auto-contours) in auto-segmentation for radiotherapy, with a focus on Online Adaptive Radiotherapy (OART). The proposed method leverages Bayesian Ordinal Classification (BOC), combined with calibrated thresholds derived from uncertainty quantification, to deliver confident QA predictions. This approach addresses key challenges in clinical auto-segmentation QA workflows such as the absence of ground truth contours, limited availability of manually labeled data, and inherent uncertainty in AI model predictions.

**Methods:** We developed a BOC model to classify the quality of auto-contours and quantify uncertainty. To enhance predictive reliability, we implemented a calibration step to determine optimal uncertainty thresholds that meet specific clinical accuracy requirements. The method was validated under three distinct data availability scenarios: absence of manual labels, limited manual labeling, and extensive manual labeling. We specifically tested our method for auto-segmented rectum contours in prostate cancer radiotherapy. Geometric surrogate labels were employed in the absence of manual labels, transfer learning was applied when manual labels were limited, and direct use of manual labels was performed when extensive labeling was available.

**Results:** The BOC model demonstrated robust performance across all data availability scenarios for confident predictions, with significant accuracy gains when pre-trained with surrogate labels and fine-tuned with limited manually labeled data. Specifically, fine-tuning the pretrained model with just 30 manually labeled cases and calibrating with 34 subjects achieved over an accuracy of over 90% against manual labels in the test dataset. Furthermore, with the calibrated uncertainty threshold, over 93% of the auto-contours' qualities were accurately predicted in over 98% of cases, significantly reducing clinician workload by minimizing unnecessary manual reviews and promptly highlighting cases needing revision.

**Conclusion:** The proposed AI-assisted auto-contour QA model effectively streamlines contouring processes, substantially reducing manual effort and improving clinical workflow efficiency in OART. By integrating uncertainty quantification, our approach enables clinicians to make rapid, informed decisions, ensuring improved patient safety and workflow reliability.

*Keywords*: Deep learning, Auto-Segmentation, Auto-Contours, Quality Assessment, Uncertainty Quantification, Online Adaptive Radiotherapy

# I. INTRODUCTION

Radiotherapy (RT) is a principal treatment for cancer, with over 50%[1] of patients receiving it as part of their regimen. Modern RT techniques focus on maximizing radiation to cancerous tissues while minimizing exposure to healthy tissues, necessitating precise segmentation of targets and organs at risk (OARs) during treatment planning. In current clinical practice, this segmentation is predominantly done manually, a process that is both time-consuming and labor-intensive.

Online adaptive radiotherapy (OART) represents a cutting-edge advancement, significantly enhancing treatment outcomes while reducing toxicity to normal tissues[2]. Despite its benefits, OART requires patients to remain immobile during treatment setup, including contour drawing and treatment planning phases, which can be uncomfortable. Streamlining the segmentation process not only improves workflow efficiency but also increases patient comfort by reducing waiting times during these critical procedures. In recent years, deep learning (DL)-based auto-segmentation has emerged as a promising solution to accelerate this process. However, some auto-generated contours (auto-contours) may not meet clinical quality standards, necessitating meticulous review by clinicians, slice-by-slice and organ-by-organ for quality assurance, which is a highly labor-intensive task. Thus, automatically assessing the quality of auto-contours is a critical downstream task for auto-segmentation, to further improve treatment efficiency, especially for time-sensitive OART.

Existing methods for assessing auto-contour quality can be categorized into three main categories:
1) Geometric-based methods[3,4]: Most geometric metrics require two contour sets (a test contour and a reference contour) to benchmark and quantify similarity. For example, the Dice Similarity Coefficient (DSC) measures the overlap between entire areas, whereas the Surface Dice Similarity Coefficient (SDSC), Hausdorff Distance (HD), Mean Surface Distance (MSD), and Added Path Length evaluate differences in contour alignment.

2) Database-driven methods: Altman et al. [5] proposed a framework to evaluate contour quality by analyzing historical data on size, shape, position, and other clinically relevant metrics, and then setting thresholds based on their distributions. However, constructing a sufficiently robust database for each organ and metric is challenging. Furthermore, the authors did not specify how these thresholds should be defined or addressed and how reliably they apply to individual clinical cases.

3) Prediction-based methods: These leverage machine learning (ML) and deep learning (DL) models to predict contour quality. Traditional ML approaches typically employ geometric metrics (e.g., DSC, SDSC, HD and MSD) as input features to predict contour quality[6, 7]. Men et al.[8] employed a deep active learning framework in a lung cancer clinical trial, generating reference contours to compute DSC and HD95 metrics for trial-submitted contours. Contours were then classified as "acceptable" or "unacceptable" based on thresholds derived from the statistical distributions of these metrics (e.g., $DSC_{test} > \mu_{DSC} - 1.96\sigma_{DSC}$ or $HD95 < \mu_{HD95} + 1.96\sigma_{HD95}$). This method depends on the accuracy and robustness of the model-generated reference contours. Chen et al.[9] trained a convolutional neural network (CNN) for breast cancer auto-contours, categorizing them as "good", "medium", or "poor" based on direct predictions from the CNN. This approach requires careful model validation to ensure consistent, clinically reliable predictions.

There are several major challenges in existing quality assessment (QA) methods. (1) Most methods rely on high-quality contours (often referred to as "ground truth" or GT) as references, which are not available in real auto-contouring scenarios. (2) Manual labeling of contour quality for model training is labor-intensive and requires specialized clinical expertise. (3) Classification-based methods fail to account for the inherent ordinal relationships among different contour quality levels and ignore the asymmetry in misclassification risks. For instance, a contour classified as "good" and usable without revision should clearly be recognized as superior to one labeled "bad", which would require further clinician review. Critically, misclassifying a poor-quality contour as acceptable can mislead clinicians and pose risks to patient care. In contrast, misclassifying a good contour as "bad" merely results in additional review time as clinicians can easily correct such errors. Given these asymmetrical risks, misclassifying a bad contour as good carries far more severe consequences than the reverse. (4) Current decision-making criteria for prediction-based methods often rely solely on QA model predictions, neglecting their inherent uncertainties. This may be insufficient, as clinical decision-making demands high contour quality and reliability of QA. Uncertainty quantification (UQ) has emerged as a prominent topic in deep learning for medical image[10], especially in classification[11-13], segmentation[14-17], dose prediction[18] and image synthesis[19-21]. However, it remains underutilized in auto-contour QA, despite its potential to improve QA reliability[12]. Our findings show that predictions with lower uncertainty consistently correlate with higher classification accuracy, reinforcing the model's reliability for clinical applications.

To address these challenges, we propose an auto-contour QA framework which integrates UQ to enhance model trustworthiness and enable the identification of reliable predictions for clinical decision-making. This study introduces a novel DL-based QA model for auto-contours, leveraging Bayesian ordinal classification (BOC) to predict contour quality and quantify uncertainty in a 2D, slice-by-slice review.

Our approach requires only CT images and corresponding auto-contours as input and can be developed using surrogate labels derived from geometric characteristics without manual labels, increasing its feasibility in clinical settings. We explore various data availability scenarios, ranging from limited manual labels with transfer learning to fully supervised training with extensive manual annotations.

To our knowledge, this is the first study to apply ordinal classification for auto-contour QA while incorporating Bayesian uncertainty quantification, tailored for DL-based ordinal classification models. In clinical practice, confident predictions with low uncertainty can highlight discrepancies between the model's assessments and clinicians' initial judgments. For example, if a high-confidence prediction of a poor-quality contour is initially deemed acceptable by clinicians, it should prompt a re-evaluation, serving as an alert. Conversely, strong agreement between the model's confident prediction and clinical assessment reinforces the contour's quality. This innovative approach promises to improve the reliability and efficiency of clinical workflows by assisting clinicians in making better-informed decisions, reducing manual review burden, and allowing them to focus on necessary contour revisions and treatment planning. By integrating UQ into DL-based auto-contour QA, our approach offers a robust solution to improve contour validation in time-sensitive radiotherapy workflows.

## II. METHODS AND MATERIALS

### A. Methods

Fig. 1 illustrates the primary workflow of our AI-assisted decision-making framework designed for clinical assessment of auto-segmented contour quality. The process begins with an AI auto-segmentation model (Step a), intended for clinical deployment, which generates the auto-contours. In Step b, these auto-contours are evaluated using our AI-based QA model, a DL-based Bayesian ordinal classification (BOC) model. The BOC model outputs predictions of contour quality along with associated uncertainty estimates. To align these uncertainties with clinical accuracy requirements, a calibration process (Step c) is conducted to establish an uncertainty threshold based on calibration data and predefined accuracy criteria. During the testing phase (Step d), only predictions with uncertainty values below this threshold are accepted as confident predictions, while those exceeding the threshold are rejected. Finally, as demonstrated in Step e, these confident predictions support clinical decision-making by determining whether corrective actions are necessary. Each step is described in detail within this section.

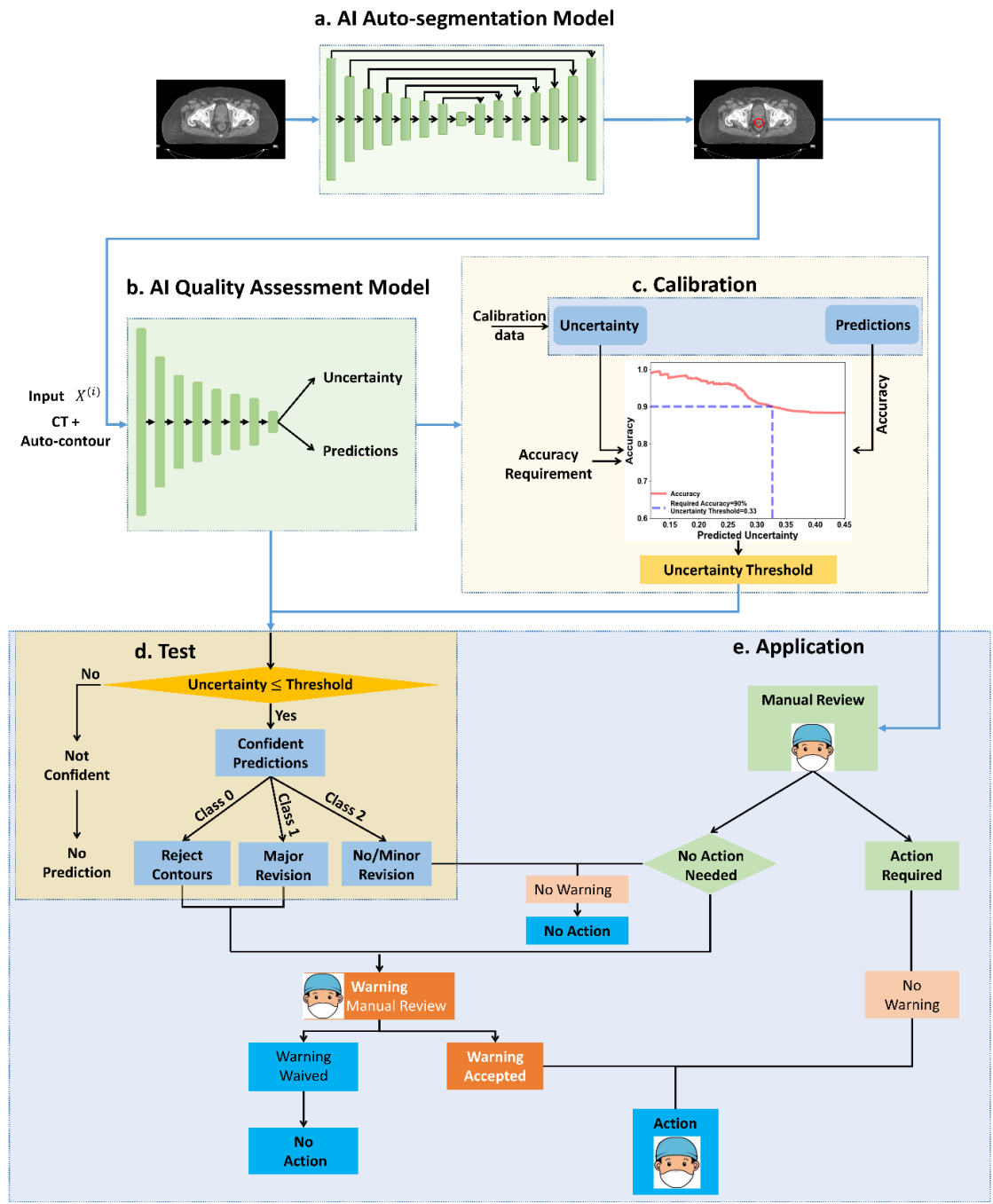

**Fig. 1 Key Workflow for AI-Assisted Decision-Making Framework.** a). An AI auto-segmentation model, which generates the auto-contours. b). AI-based quality assessment model to evaluate these auto-contours, a deep learning Bayesian ordinal classification (BOC) model which outputs contour quality along with associated uncertainty estimates. c) Calibration, this process is conducted to establish an uncertainty threshold based on calibration data and predefined accuracy criteria. d) Test, only predictions with uncertainty values below this threshold are accepted as confident predictions for next step. e) Application, these confident predictions support clinical decision-making by determining whether corrective actions are necessary.

## 1) Deep Learning-based Bayesian Ordinal Classification (BOC) Model

Let $\mathcal{X}$ denote the input space consisting of all 2D CT slices together with their associated auto-contours. Let $\mathcal{Y}$ denote the label space of ordinal quality levels, $\mathcal{Y} = \{r_0, r_1, \ldots, r_k, \ldots r_{K-1}\}$ with $r_0 = 0, r_k = k, k = 1,2,\ldots,K-1$, and $r_{K-1} > \cdots > r_0$. Given a training dataset $D = \{X^{(i)}, y^{(i)}\}_{i=1}^{N}$, each $X^{(i)} \in \mathcal{X}$ represents the $i$-th 2D slice with its auto-contour, $y^{(i)} \in \mathcal{Y}$ is the corresponding auto-contour quality label, and $N$ is the total number of training samples, $i = 1,2,\ldots N$.

To classify these ordinal labels, we first use ResNet34 with network parameters W to extract features from each input $X^{(i)}$. In the output end, conventional $K$-class nominal classification requires $K$ output units, while our BOC framework uses $K$-1 binary classifiers. Each contour quality label $y^{(i)}$ is coded into $K$-1 binary labels, and $y_k^{(i)} \in \{0,1\}, k = 1,2,\cdots,K-1$, where $y_k^{(i)} = 1$ means $y^{(i)} \geq r_k$, and $y^{(i)} = \sum_{k=1}^{K-1} y_k^{(i)}$. For example, with $K = 3$: if $y^{(i)} = 2$, then $y_1^{(i)} = 1$ and $y_2^{(i)} = 1$, it is coded as (1, 1); similarly, $y^{(i)} = 1$ is coded as (1,0), and $y^{(i)} = 0$ as (0,0). Each binary classifier outputs a probability $f_k(X^{(i)}|W)$ to model the conditional probability $p(y_k^{(i)} = 1|y_{k-1}^{(i)} = 1, X^{(i)})$ for label $y_k^{(i)}$. That is,
$$p(y^{(i)} \geq r_k | y^{(i)} \geq r_{k-1}, X^{(i)}) = p(y_k^{(i)} = 1|y_{k-1}^{(i)} = 1, X^{(i)}) = f_k(X^{(i)}|W).$$
When $k=1$, this conditional probability $f_1(X^{(i)}|W)$ becomes the marginal probability $p(y_1^{(i)} = 1|X^{(i)})$. And the marginal probability for each label follows the chain rule:
$$p(y^{(i)} \geq r_k|X^{(i)}) = \prod_{j=1}^{k} f_j(X^{(i)}|W).$$
In our auto-contour QA context, we use three ordinal classes (i.e., $K = 3$): 'Not Acceptable' (or 'Rejection', $r_0 = 0$), 'Major Revision' ($r_1 = 1$), and 'No/Minor Revision' ($r_2 = 2$). Thus, $y^{(i)}$ corresponds to two binary labels $(y_1^{(i)}, y_2^{(i)})$. According to the BOC setting, $y_1^{(i)}$ and $y_2^{(i)}$ can be viewed as Bernoulli variables with probabilities $f_1(X^{(i)}|W)$ and $f_1(X^{(i)}|W)f_2(X^{(i)}|W)$, respectively. But different from nominal classification, they follow an ordinal structure, if $y_1^{(i)}$ is 0, then $y_2^{(i)}$ must be 0, indicating that no higher class is reached. If $y_1^{(i)}$ is 1, then $y_2^{(i)}$ is determined by $f_2(X^{(i)}|W)$. Consequently, $p(y^{(i)} \geq r_1|X^{(i)}) = f_1(X^{(i)}|W)$, $p(y^{(i)} \geq r_2|X^{(i)}) = f_1(X^{(i)}|W)f_2(X^{(i)}|W)$.

In our BOC setting, the network weights (including bias terms) are treated as random variables denoted by $W \in \Omega$, where $\Omega$ is the parameter space. We assume a prior distribution $p(W)$ for $W$ over $\Omega$. Given data $= (X, y)$, the posterior distribution of $W$ is:
$$p(W|D) = \frac{p(y|W,X)p(W)}{p(y|X)},$$
where, $p(y|W,X) = \prod_{i=1}^{N} \prod_{k=1}^{K-1} f_k(X^{(i)}|W)^{\mathbb{I}\{y^{(i)} \geq r_k\}}(1 - f_k(X^{(i)}|W))^{\mathbb{I}\{y^{(i)} < r_k\}}$.

Then the posterior distribution is given by:
$$p(W|D) \propto \prod_{i=1}^{N} \prod_{k=1}^{K-1} f_k(X^{(i)}|W)^{\mathbb{I}\{y^{(i)} \geq r_k\}}(1 - f_k(X^{(i)}|W))^{\mathbb{I}\{y^{(i)} < r_k\}} p(W).$$

For a new input $X^*$ with an unknown label $y^*$, the posterior predictive distribution is:
$$p(y^*|X^*, D) = \int_{\Omega} p(y^*|X^*, W)p(W|D))dW.$$
And the conditional posterior predictive distributions for $y_1^*$ and $y_2^*$ are:
$$p(y_1^* = 1|X^*, D) = \int_{\Omega} f_1(X^*|W)p(W|D)dW$$
$$p(y_2^* = 1|y_1^* = 1, X^*, D) = \int_{\Omega} f_2(X^*|W)p(W|D)dW$$

The final quality label is $y^* = y_1^* + y_2^*$.

The goal of this BOC model is to learn a mapping $\mathcal{X} \to \mathcal{Y}$ that optimizes the posterior distribution $p(W|D)$. Since the posterior distribution $p(W|D)$ is intractable, we employ variational inference, commonly used for posterior approximation. We first define the variational distribution as $q_\theta(w) \in \mathcal{Q}$, where $\theta$ is the variational parameter with $\theta \in \Theta$, $\Theta$ is the variational parameter space, $\mathcal{Q}$ is an approximate density family. Then we minimize the Kullback-Leibler (KL) divergence, measuring the distance between $q_\theta(w)$ and the target $p(W|D)$ to obtain the optimal variation distribution $q_\theta^*(W)$:

$$q_\theta^*(W) = \underset{q_\theta(W) \in \mathcal{Q}}{\operatorname{argmin}} \operatorname{KL}(q_\theta(W) || p(W|D))$$

$$\operatorname{KL}(q_\theta(W) || p(W|D)) = -\int q_\theta(W) \log p(y|W,X) dW + \operatorname{KL}(q_\theta(W) || p(W)) + \log p(y|X))$$

$$= -\int q_\theta(W) \log p(y|W,X) dW + \operatorname{KL}(q_\theta(W) || p(W)) + \text{Constant}$$

Thus, to minimize the KL-divergence is to minimize the negative evidence lower bounder (ELBO):

$$-\operatorname{ELBO} = -\int q_\theta(W) \log p(y|W,X) dW + \operatorname{KL}(q_\theta(W) || p(W))$$

Since Gal et al.[22] proved that a neural network trained with dropout incorporating stochastic regularization techniques is a Bayesian neural network, we trained our BOC model with dropout layers to get the an optimal variational approximation $q_\theta^*(W)$. We adopted the training procedure from Shi et al. [23] to estimate the conditional probabilities.

### 2) Uncertainty Quantification Method
#### a) Uncertainty Estimated from the BOC Model

To quantify prediction uncertainty, we consider the variance of $y^*$,

$$Var(y^*) = Var(\sum_{k=1}^{K-1} y_k^*) = Var(y_1^* + y_2^*).$$

The first two central moments of $y^*$ with respect to the distribution $p(y^*|X^*, W)$ can be given by:

$$\mathbb{E}_{p(y^*|X^*,W)}(y^*) = \mathbb{E}_{p(y^*|X^*,W)}(y_1^*) + \mathbb{E}_{p(y^*|X^*,W)}(y_2^*)$$

$$= \sum_{y_1^*} \{y_1^* p(y_1^*|X^*, W)\} + \sum_{y_2^*} \{y_2^* p(y_2^*|X^*, W)\}$$

$$= f_1(X^*|W) + f_1(X^*|W) f_2(X^*|W)$$

$$Var_{p(y^*|X^*,W)}(y^*) = Var_{p(y^*|X^*,W)}(y_1^*) + Var_{p(y^*|X^*,W)}(y_2^*) + 2Cov_{p(y^*|X^*,W)}(y_1^*, y_2^*)$$

$$= \sum_{y_1^*} \{y_1^{*2} p(y_1^*|X^*, W)\} - f_1(X^*|W)^2 + \sum_{y_2^*} \{y_2^{*2} p(y_2^*|X^*, W) - f_1(X^*|W)^2 f_2(X^*|W)^2$$

$$+ 2 \sum_{y_1^* y_2^*} \{y_1^* y_2^* p(y^*|X^*, W)\} - 2 f_1(X^*|W)^2 f_2(X^*|W)$$

$$= f_1(X^*|W) + 3 f_1(X^*|W) f_2(X^*|W) - f_1(X^*|W)^2 (1 + f_2(X^*|W))^2$$

The first two central moments of $y^*$ with respect to the posterior predictive distribution $p(y^*|X^*, D)$ can be given by:

$$\mathbb{E}_{p(y^*|X^*,D)}(y^*) = \int_\Omega f_1(X^*|W) p(W|D) dW + \int_\Omega f_1(X^*|W) f_2(X^*|W) p(W|D) dW \quad \text{(2-1)}$$

$$Var_{p(y^*|X^*,D)}(y^*) = \mathbb{E}_{p(W|D)} Var_{p(y^*|X^*,W)}(y^*|W) + Var_{p(W|D)} \mathbb{E}_{p(y^*|X^*,W)}(y^*|W)$$

$$= \int_\Omega \{f_1(X^*|W) + 3 f_1(X^*|W) f_2(X^*|W) - f_1(X^*|W)^2 (1 + f_2(X^*|W))^2\} p(W|D) dW + \int_\Omega \{f_1(X^*|W) +$$

$$f_1(X^*|W) f_2(X^*|W)\}^2 p(W|D) dW - \{\int_\Omega f_1(X^*|W) p(W|D) dW + \int_\Omega f_1(X^*|W) f_2(X^*|W) p(W|D) dW\}^2 \quad \text{(2-2)}$$

Then we used the variational distribution $q_\theta(W)$ for $p(W|D)$ to approximate the posterior predictive distribution:

$$q_\theta(y^*|X^*, D) = \int_\Omega p(y^*|X^*, W) q_\theta(W) dW$$

for all $\theta \in \Theta$. Following this, we can get the variational approximation of $Var_{p(y^*|X^*, D)}(y^*)$ by replacing $p(W|D)$ in formula (2-2) by $q_\theta(W)$.

To avoid the high-dimensional integration during the inference phase, a Monte Carlo (MC) estimator is used to draw inference, where the set $\{W_t\}_{t=1}^T$ is randomly drawn from $q_\theta(W)$.

$$\hat{p}_\theta(y^*|X^*, D) = \frac{1}{T}\sum_{t=1}^T p(y^*|X^*, W_t) \xrightarrow{T \to \infty} \int_\Omega p(y^*|X^*, W) q_\theta(W) dW = q_\theta(y^*|X^*)$$

When the optimized variational parameter is obtained, expressed by $\hat{\theta}$ and the corresponding estimate for $W$ is $\hat{W}$, then we have:

$$\hat{p}_{\hat{\theta}}(y^*|X^*, D) = \frac{1}{T}\sum_{t=1}^T p(y^*|X^*, \hat{W}_t)$$

and

$$\hat{p}_{\hat{\theta}}(y^* \geq r_k | y^* \geq r_{k-1}, X^*, D) = \hat{p}_{\hat{\theta}}(y_k^* = 1 | y_{k-1}^* = 1, X^*)$$
$$= \frac{1}{T}\sum_{t=1}^T p(y_k^* = 1 | y_{k-1}^* = 1, X^*, \hat{W}_t)$$
$$= \frac{1}{T}\sum_{t=1}^T \hat{f}_k(X^*|\hat{W}_t)$$

Then the MC estimators for the two probabilities are given by:

$$\hat{p}_{\hat{\theta}}(y_1^* = 1|X^*, D) = \frac{1}{T}\sum_{t=1}^T \hat{f}_1(X^*|\hat{W}_t)$$

$$\hat{p}_{\hat{\theta}}(y_2^* = 1|y_1^* = 1, X^*, D) = \frac{1}{T}\sum_{t=1}^T \hat{f}_2(X^*|\hat{W}_t)$$

such that the MC estimators for the first two moments can be derived as follows:

$$\hat{\mathbb{E}}_{p(y^*|X^*, D)}(y^*) = \frac{1}{T}\sum_{t=1}^T \hat{f}_1(X^*|\hat{W}_t) + \frac{1}{T}\sum_{t=1}^T \hat{f}_1(X^*|\hat{W}_t)\hat{f}_2(X^*|\hat{W}_t) \quad (2\text{-}3)$$

$$\hat{Var}_{p(y^*|X^*, D)}(y^*) = \frac{1}{T}\sum_{t=1}^T \{\hat{f}_1(X^*|\hat{W}_t) + 3\hat{f}_1(X^*|\hat{W}_t)\hat{f}_2(X^*|\hat{W}_t) - \hat{f}_1(X^*|\hat{W}_t)^2(1+\hat{f}_2(X^*|\hat{W}_t))^2\}$$
$$+ \frac{1}{T}\sum_{t=1}^T \{\hat{f}_1(X^*|\hat{W}_t) + \hat{f}_1(X^*|\hat{W}_t)\hat{f}_2(X^*|\hat{W}_t)\}^2 \quad (2\text{-}4)$$
$$- \{\frac{1}{T}\sum_{t=1}^T \{\hat{f}_1(X^*|\hat{W}_t) + \hat{f}_1(X^*|\hat{W}_t)\hat{f}_2(X^*|\hat{W}_t)\}\}^2$$

$\hat{\mathbb{E}}_{p(y^*|X^*, D)}(y^*)$ is the estimated mean of the contour quality, and $\hat{Var}_{p(y^*|X^*, D)}(y^*)$ is the corresponding estimated uncertainty. And the predicted label for contour quality is:

$\hat{y}^* = \sum_{k=1}^{K-1} \hat{y}_k^*$, where $\hat{y}_k^* = \mathbb{I}\{\hat{p}_{\hat{\theta}}(y^* \geq r_k | y^* \geq r_{k-1}, X^*) > 0.5\}$. (2-5)

#### b) Assessment of Model Estimated Uncertainty against Manual Label Uncertainty

A major challenge in uncertainty quantification for DL is that there is no exact 'true' uncertainty value to compare against. This makes it difficult to evaluate uncertainty quantification methods. To address this, we estimated the uncertainty associated with manual labels provided by multiple clinicians (i.e. inter-observer variation) for the purpose of evaluating our model's uncertainty estimates. Three clinicians independently reviewed and labeled the test samples. When a contour's quality is highly uncertain (i.e., its category is difficult to determine), clinicians are more likely to assign different labels to the same contour. Therefore, an effective uncertainty quantification method should exhibit a strong correlation between the uncertainty in clinicians' assessments and the model's uncertainty estimates.

We employed an entropy-based method[22] to quantify the uncertainty of manual labels:

$$H\{p(y^{(i)})\} = -\sum_j p_j(y^{(i)} = j)\log(p_j(y^{(i)} = j))$$

where $i$ denotes the $i$-th contour, and $p_j(y^{(i)} = j)$ is the probability that the $i$-th contour is assigned to quality class $j$ by clinicians, with $j \in \{0,1,2\}$ in our study. Each clinician's label was given equal weight in determining the class of a contour, leading to possible probability values of $p_j \in \{0, \frac{1}{3}, \frac{2}{3}, 1\}$. To ensure numerical stability, we follow the convention that $0 \log 0 = 0$ in the entropy calculation, thereby avoiding undefined values when $p_j = 0$.

High entropy indicates disagreement among clinicians, reflecting higher uncertainty, while low entropy indicates consensus, denoting lower uncertainty. Since we relied on labels from only three clinicians, the entropy-based uncertainty measure yielded three discrete uncertainty values for the manual labels: 1) full agreement (all three clinicians assigned the same label), 2) partial agreement (two clinicians assigned the same label while the third differed), and 3) complete disagreement (each clinician assigned a different label).

To address our model's ability to estimate uncertainty, we grouped the test data according to these three values of manual-label derived uncertainty. For each group, we calculated the mean predicted uncertainty from our BOC model and analyzed the correlation between the manual-label uncertainty values and the corresponding mean model-predicted uncertainty. A strong positive correlation would indicate that the model effectively captures the level of uncertainty reflected in clinician assessments, supporting its reliability in uncertainty estimation.

### 3) AI-Assisted Decision-making Strategies for Application
#### a) Calibration

Our BOC model integrated both estimated uncertainty and predicted quality in decision-making processes for auto-contour QA. We hypothesized that effective uncertainty quantification would demonstrate a direct correlation between decreased uncertainty and improved model performance.

To quantify this relationship, we incorporated a calibration step (Fig. 1c), where calibration results were sorted in the ascending order of uncertainty, and cumulative accuracies were computed. A reliable uncertainty quantification method should produce an "accuracy vs. uncertainty" curve with a decreasing trend.

Using this curve, we established an uncertainty threshold aligned with clinical accuracy requirements. For example, if a clinical application required at least 90% accuracy, we identified this accuracy level on the vertical axis and determined the corresponding uncertainty value on the horizontal axis as the threshold for future testing.

Assuming the test data followed the same distribution as the calibration data, we anticipated predictions with uncertainty below this threshold to achieve the required 90% accuracy. Predictions meeting this criterion were deemed sufficiently reliable for confident predictions from our BOC model, ensuring both model reliability and applicability in future applications.

### b) Test and Application

Fig. 1d illustrates that during testing, predictions of contour quality with uncertainty below the established threshold are accepted, while those exceeding this threshold are rejected. This selective acceptance ensures that only confident predictions are presented to clinicians, enhancing the model's clinical utility by avoiding unreliable predictions.

The application section in Fig. 1e details how these confident predictions are applied in a clinical context to assist in decision-making. This approach is to streamline the auto-contour quality evaluation process and improve the reliability of clinical decisions, particularly when no contour revision is needed. Our strategies include:

**No Warning:** If clinicians and our model agree on the contour quality, or if clinicians independently determine that significant adjustments are required, regardless of any discrepancy with the model, no warning is triggered. Clinicians can then proceed accordingly, accepting the auto-contours, making revisions, or rejecting and redrawing contours as needed.

**Warning:** If our model predicts that significant adjustments are necessary ('Major Revision' or 'Not Acceptable', corresponding to Class 1 or 0), but clinicians initially determine that no revisions are needed, the system triggers a warning. This serves as a prompt for clinicians to critically reevaluate the auto-contours. After reassessment, clinicians can dismiss the warning and make an informed decision to accept, revise, or reject and redraw contours.

### c) Model Developing for Different Scenarios

Training DL models typically requires a large amount of labeled data. Manually labeling contour quality is time-intensive and demands significant effort from highly trained clinicians, whose availability is often constrained by intense clinical responsibilities. This study investigates the adaptability of our proposed methods under real-world conditions, with a particular focus on the feasibility and availability of manual labels.

When manual labels are unavailable, we use geometric metrics based surrogate labels to train and evaluate our BOC model. In scenarios where labeling a small dataset is feasible, human-labeled data are used to fine-tune our pre-trained model. We explore transfer learning from a model initially trained with surrogate labels, investigating various sample sizes to determine the minimum number required for effective transfer learning. When sufficient manual labels are available, they are directly used to train the BOC model from scratch. For both transfer learning and training from scratch with manual labels, we use additional manual labels to evaluate model performance.

The comprehensive workflow is illustrated in Fig. 2 without the b2 part. The process begins with training a 3D auto-segmentation model, whose predictions are used to extract 2D slices for auto-contour QA. These 2D auto-contours, along with CT images and labels, either surrogate or manual, are used to develop the DL-based BOC model for assessing contour quality. The overall workflow remains the same regardless of whether training is performed with surrogate labels or manual labels from scratch, and testing labels always match the type of labels used for training.

In scenarios where limited manual labels are accessible, an additional transfer learning step is introduced to enhance the model initially developed with surrogate labels, as detailed in Fig. 2 with the b2 part. When sufficient manual labels are available, they are directly used to train the BOC model. Regardless of the training approach, the subsequent phases—calibration, testing, and application—are consistent across all scenarios, as previously discussed.

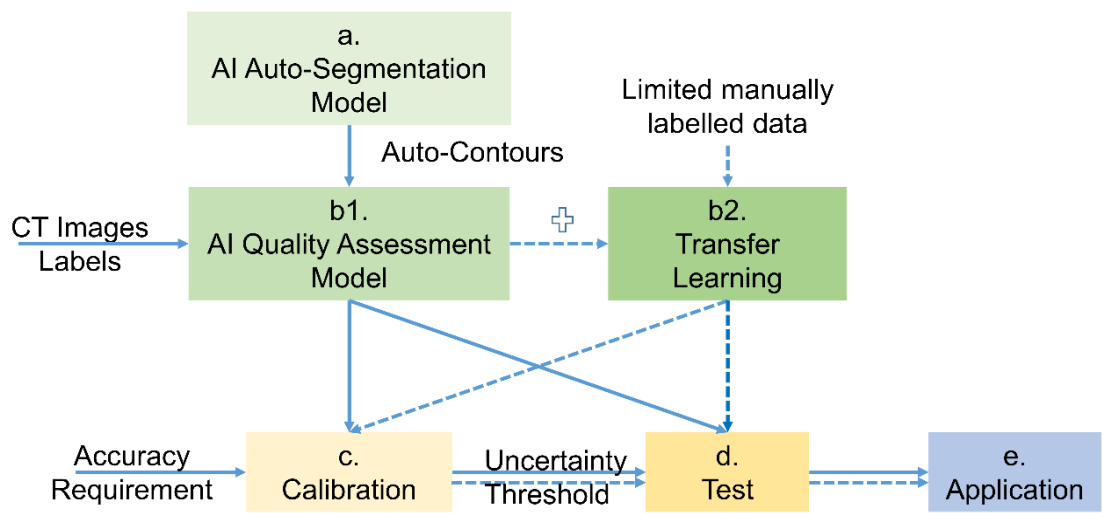

**Fig. 2 Main Steps for Model Developing in Different Scenarios.** Models trained with different strategies follow slightly different workflows. Those trained from scratch using surrogate or manual labels follow steps a–b1–c–d–e, while models trained via transfer learning follow steps a–b1–b2–c–d–e.

### B. Materials

#### 1) Data

##### a) Image Data

In this single-institution study approved by our institutional review board (IRB), we curated a cohort of 1,328 prostate cancer patients from University of Texas Southwestern Medical Center, each with at least one cleaned GT contour for the rectum, bladder or prostate. The CT scans from these patients had slice thickness ranging from 1.5 to 3.0 mm, resolution of 0.9 to 1.5 mm, image size of 512x512 pixels, and 90 to 290 slices per scan.

We randomly selected 500 subjects to develop our in-house DL-based auto-segmentation model, with 450 subjects for training and 50 for validation. For the remaining 828 subjects, we generated auto-contours using the validated auto-segmentation model for further development of our QA framework.

Given the increased complexity in rectum segmentation compared to other OARs such as the bladder and femurs, in prostate cancer RT, we selected rectum as the primary focus of this study to evaluate the feasibility of our auto-contour QA approach.

Among the 828 subjects, we identified 784 subjects with both GT and predicted contours for the rectum. We then divided them into different groups for model development and application: (i) 292 subjects (11,934 slices) for training, (ii) 34 subjects (1,441 slices) for validation, (iii) 297 subjects (12118 slices) for calibration, and (iv) 114 subjects (4,614 slices) for model-developing under other scenarios to explore transfer learning and train from scratch with manual labels. An extra dataset containing (v) 47 subjects (1,922 slices) was saved for testing all models.

### b) *Auto-contour quality labels*

**Contour Quality Ranking System**

Auto-contour quality is commonly rated using 3-point[9, 24, 25] or 5-point[3, 26] scale ranking systems. In our study, we adopted a 3-point scale to quantify the contour quality based on the following clinical 'acceptability' criteria:

- **Class 2 (No/Minor Revision):** Clinically acceptable contours, though some clinicians may suggest minor edits due to inter-observer variability.
- **Class 1 (Major Revision):** Generally good contours requiring significant modifications for clinical use, but still usable as a starting point to save time.
- **Class 0 (Not Acceptable):** Inadequate contours that cannot serve as a starting point and must be fully redrawn by clinicians.

**Surrogate Labels**

Manually labeling the quality of auto-contours is both labor-intensive and time-consuming, making it challenging to accumulate a sufficiently large dataset of clinician-labeled samples for training DL-based QA models. However, research shows significant organ-specific correlations between various metrics and manually ranked contour quality for prostate, rectum, bladder, and seminal vesicles, with correlation coefficients ranging from 0.25 to 0.59[26]. Studies also demonstrate that combining multiple metrics provides a more robust assessment than relying solely on DSC or HD95 for contour quality evaluation. [8], [26], [27], [28]

To address the scarcity of clinician-labeled data for training, our study introduces a surrogate labeling system for auto-contour quality that integrates multiple geometric metrics. DSC correlates highly with contour size; according to Men et al.[8], smaller contours tend to exhibit lower DSC scores but also lower HDs. In contrast, SDSC[29] shows greater independence from contour size compared to DSC. Thus, our model incorporated the three geometric metrics: DSC for measuring overlap index and SDSC for surface agreement, along with HD95 for boundary distance.

For clinical application, we need to define specific thresholds for each metric across different OARs. Auto-contours are categorized into different quality classes based on these thresholds, with the final classification determined by the highest value among $\{r_{DSC}, r_{SDSC}, r_{HD95}\}$. Table 1 outlines the classification thresholds for rectum contours based on geometric metrics.

Table 1 Labeling Strategy for Surrogate Labels Based on Geometric Metrics

| Class | 2 (No/Minor Revision) | 1 (Major Revision) | 0 (Not Acceptable) |
|---|---|---|---|
| $r_{DSC}$ threshold | $DSC \in [0.9, 1]$ | $DSC \in [0.7, 0.9)$ | $DSC \in [0, 0.7)$ |
| $r_{SDSC}$ threshold | $SDSC \in [0.9, 1]$ | $SDSC \in [0.7, 0.9)$ | $SDSC \in [0, 0.7)$ |
| $r_{HD95}$ threshold | $HD95 \in [0, 2.5]$ | $HD95 \in (2.5, 6]$ | $HD95 \in (6, +\infty)$ |
| Final Class Rule | \multicolumn{3}{c}{Final Class = $max(r_{DSC}, r_{SDSC}, r_{HD95})$} | |

**Manual Labels**

Three clinicians independently rated the quality of auto-contours in group (v). Each clinician reviewed all 1922 slices from the 47 subjects. The consensus quality class for each auto-contour was determined by majority vote. In instances where the opinions of the three clinicians diverged, the contour quality was assigned a manual label of Class 1 ('Major Revision'). These manual labels were then used to benchmark the performance of our automated QA model. Additionally, one clinician labeled an expanded set of 114 subjects (4614 slices) for use in studies exploring transfer learning and training with manual labels from scratch for our BOC model development. Contour quality was assessed using the previously mentioned 3-point ranking system, based on clinicians' expertise and understanding of clinical acceptability.

### c) Data Augmentation for Surrogate Labels

Data augmentation is widely used in medical image analysis to address data scarcity and improve model performance and generalizability[30]. In this study, data augmentation refers to the process of artificially generating diverse samples by applying transformations to existing contours.

We employed data augmentation to improve our BOC model trained with surrogate labels. We used clinical accepted contours as references and modified them using the 'Rand2DElastic' function from MONAI library to generate additional contours. This function applied various transformations which includes affine adjustments, spacing alterations, rotations, scaling, and translations, to simulate the variability introduced by different auto-segmentation models. The generated contours were then assigned surrogated labels based on their geometric agreement with the clinically accepted contours, evaluated using DSC, SDSC and HD95 as previously described. This approach enabled the creation of a diverse dataset of synthetic contours, mirroring the variations expected from different DL-based auto-segmentation models.

### 2) Experimental Design
### a) DL- based auto-segmentation model

We trained an in-house 3D auto-segmentation model based on the open-source MONAI U-Net to generate auto-contours of the prostate, rectum and bladder for the downstream QA task. The model was trained using the Adam optimizer with default hyperparameters ($\beta_1 = 0.9$, and $\beta_2 = 0.999$) over $1 \times 10^5$ iterations, leveraging the dice loss function. We initiated the learning rate at $1 \times 10^{-4}$, reducing it to $1 \times 10^{-5}$ at the $4 \times 10^4 th$ iteration and $1 \times 10^{-6}$ at the $8 \times 10^4 th$ iteration, respectively. We set the batch size to one. The data was randomly allocated to train, validation, and test data sets with 450, 50 and 828 subjects, respectively. We then used the predicted 3D contours from the trained auto-segmentation model to extract 2D contours for each slice of each organ.

### b) BOC model

We used ResNet-34 as the baseline architecture for our BOC model, which classifies contour quality by distinguishing 'bad' contours (Class 0) from 'usable' contours (Class 1 and Class 2), while further identifying Class 2 contours within the 'usable' group. Each input consisted of a CT image and an auto-contour at a 512x512 resolution. The model outputs two probabilities: $\hat{p}(y_1^* = 1)$, and the conditional probability for $\hat{p}(y_2^* = 1|y_1^* = 1)$ from which the final binary outputs are obtained according to equation (2-5).

To implement a variational approximation for a Bayesian Neural Network, we modified ResNet-34 by adding dropout layers (drop rate=0.1) after each activation layer. The network was trained using the Adam optimizer with default hyperparameters ($\beta_1 = 0.9$, and $\beta_2 = 0.999$) for 600 epochs (more than 447,000 iterations) with a default weight decay parameter (1e-4). We used the CORN[23] loss function and applied a multi-step learning rate scheduler, reducing the rate to 20% of its previous value at specified epochs (50, 100, 200, 300, 400). The batch size was set to16 throughout.

With manual labels present for 4614 slices from the 114 subjects in group (iv), we explored both transfer learning and training from scratch by varying the size of labeled data (10-80 subjects) and using 5-fold cross-validation, while the remaining were reserved for calibration. This approach helped us assess the feasibility of manual-label-based training and identify the sample size needed with and without transfer learning. Specifically, for transfer learning, we fine-tuned the pre-trained BOC model with varying initial learning rates for successive layers ($1\times10^{-6}$ for the first two grouped layers, $1\times10^{-5}$ for the intermediate two layers and $1\times10^{-4}$ for the last two layers), applying a learning rate scheduler that reduced each rate to 20% of its previous value every 100 epochs.

### c) MC dropout for predictions and uncertainty quantification

During calibration and testing, we used the MC dropout[31] to get both predictions and uncertainty estimates. Specifically, we performed 20 forward passes per sample (T=20), following the formulas in Section II-A-2). This choice of T aligns with findings by Kwon et al.'s[15], who demonstrated that T=20 provides a practical balance between computational efficiency and the precision of uncertainty estimation.

## III. RESULTS

### A. Model Developed with Surrogate Labels.

#### 1) Uncertainty Guided Decision Making

**Overall Performance**

In testing samples, the model trained with surrogate labels achieved an overall accuracy of 70.8%, which increased to 79.2% with data augmentation. As shown in Table 2, the ROC AUC values increased from 0.821 to 0.932 and from 0.778 to 0.808 for predicting Class 2 and Class 1, respectively. The model with data augmentation also achieved ROC AUC scores of 0.988 for Class 0 and 0.932 for Class 2, indicating that training with surrogate labels effectively enables the BOC model to distinguish both very good and very poor auto-contours.

Table 2 Results for Model Trained with Surrogate Labels

|  | No Augmentation | | | Data Augmentation | | |
|---|---|---|---|---|---|---|
|  | Label-wise | | | Label-wise | | |
|  | Class0 | Class1 | Class2 | Class0 | Class1 | Class 2 |
| Weighted Precision | 0.947 | 0.720 | 0.754 | 0.942 | 0.861 | 0.854 |
| Weighted Recall | 0.951 | 0.713 | 0.754 | 0.938 | 0.813 | 0.852 |
| Weighted F1 | 0.943 | 0.716 | 0.754 | 0.936 | 0.831 | 0.852 |
| **AUC** | **0.909** | **0.778** | **0.821** | **0.988** | **0.808** | **0.932** |
| **Overall Accuracy** | 70.8% | | | 79.2% | | |

**Uncertainty Guided Predictions**

During calibration, the predicted uncertainties were sorted and binned (Fig. 3, panels a1 and b1) to illustrate how accuracy declines as uncertainty increases. Based on this relationship, we established an uncertainty threshold ($\tau$) to meet a predefined accuracy requirement. For example, to maintain 90% accuracy for clinical use, $\tau = 0.13$ and $\tau = 0.24$ were established at the intersection of the 90% accuracy horizontal line with the accuracy vs. uncertainty curve in the settings without and with data augmentation, respectively (Fig. 3, panels a2 and b2). These threshold values were subsequently applied for downstream testing and application. Table 3 summarizes different predefined accuracy requirements (90%, 85%, and 80%) and their corresponding uncertainty thresholds for models trained with and without data augmentation.

During testing (Fig. 3, panels a3 and b3), the negative correlation between accuracy and uncertainty persisted in both settings. Notably, 60.5% of the cases exhibited uncertainties below $\tau$, achieving 93.2% accuracy in the setting with data augmentation (Fig. 3, panel b4) while only 20.0% of the cases exhibited uncertainties below $\tau$ achieving 91.9% accuracy in the setting without data augmentation (Fig. 3, panel a4). If the required accuracy was lowered to 80%, then 90.4% of the cases attained an overall accuracy of 83.3% (Table 3). These findings confirm the efficacy of our BOC model, trained with surrogate labels and data augmentation, in producing reliable auto-contour quality predictions for clinical application. Moreover, this uncertainty-guided approach can be flexibly adapted to meet different clinical accuracy requirements, thereby supporting clinicians in their decision-making processes.

Table 3 Uncertainty Guided Accuracy for Model Trained with Surrogate Labels

|  | No Augmentation | | | Data Augmentation | | |
| --- | --- | --- | --- | --- | --- | --- |
| Accuracy Requirement | 90.0% | 85.0% | 80.0% | **90.0%** | **85.0%** | **80.0%** |
| Uncertainty Threshold $\tau$ | 0.13 | 0.21 | 0.28 | **0.24** | **0.28** | **0.34** |
| Surrogate Label Accuracy | 91.9% | 85.4% | 81.5% | **93.2%** | **88.9%** | **83.3%** |
| Case proportion | 20.0% | 37.0% | 49.4% | **60.5%** | **74.4%** | **90.4%** |

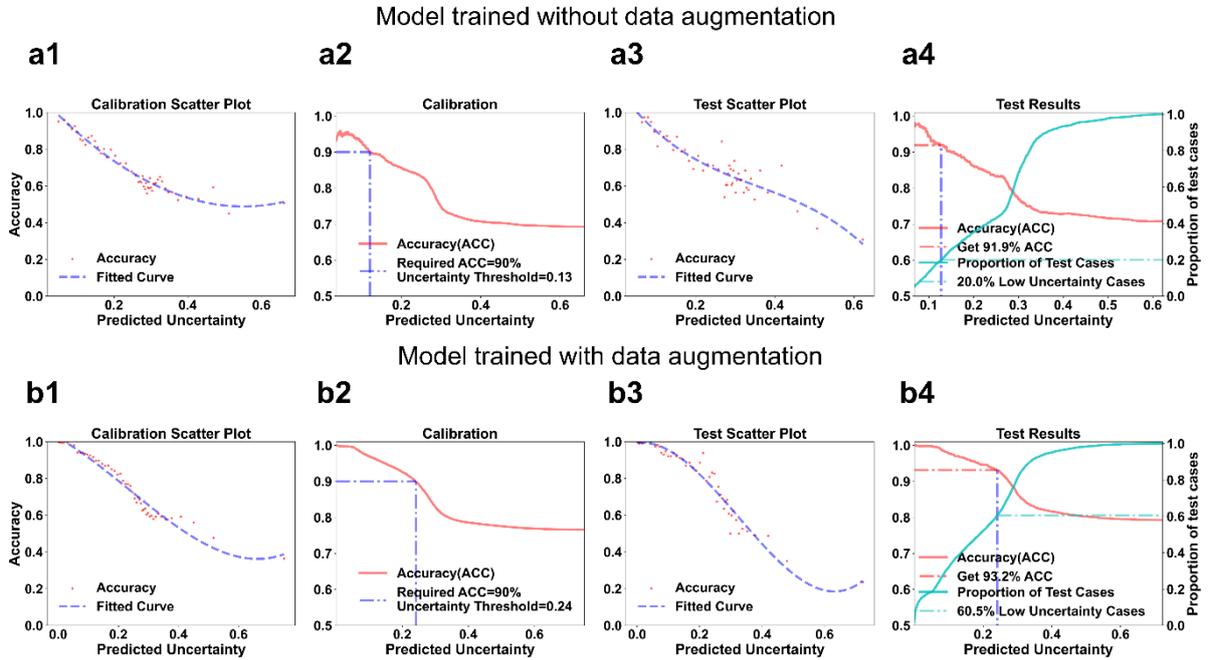

**Fig. 3 Accuracy vs. Uncertainty on Calibration and Test Data.** a1-a4) Train the BOC model with surrogate labels only (W/O augmentation), a1 and a3 show accuracy decreases with increasing uncertainty for both calibration and test data sets. a2 illustrates how to get the uncertainty threshold based on the 'accuracy vs. uncertainty' curve. a4 exhibits utilizing the uncertainty threshold 0.13 from calibration, we get 91.9% accuracy for 20.0% cases with uncertainty below the threshold.
b1-b4) Train the BOC model with surrogate labels only (W/ augmentation), each subplot shows similar information but different uncertainty and accuracy.

### 2) Results Validated with Clinician-Generated Labels
**Assessment of Predicted and Manual Label Uncertainty**

As detailed in Section II.A.2b, we grouped the test data by the uncertainty derived from manual labels from three clinicians and compared these values to the uncertainty estimates from our BOC model. Fig. 4a-c demonstrates a strong positive correlation between the uncertainties quantified by our method and those derived from manual labels, with a Spearman correlation coefficient of 1 across all three models trained using different strategies. Furthermore, cases that the BOC model identified as having higher uncertainty were similarly challenging for clinicians, demonstrating the effectiveness of our uncertainty estimation. Detailed information on the clinicians' evaluations of the test data set is provided in the Appendix.

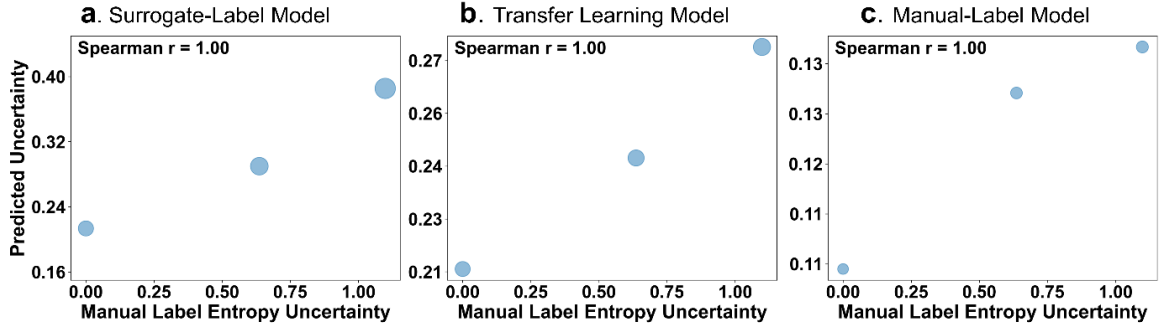

**Fig. 4 Predicted Uncertainty vs. Manual Label Uncertainty.** a–c show a consistent positive correlation between the uncertainties estimated by our method and those derived from manual labels, across models trained using different strategies.

**Assessment of Accuracy with Manual labels**

Our models, initially trained with surrogate labels derived from geometric characteristics of auto-contours, were evaluated against manually labeled clinical data. Although discrepancies between geometric and manual labeling are well-documented [8, 13-15], the two labelling strategies exhibit a strong correlation. When directly compared with manual labels, our model trained with surrogate labels achieved an accuracy of approximately 73.3% when using data augmentation. Furthermore, incorporating an uncertainty threshold to enhance prediction confidence substantially improved alignment with manual labels. As detailed in Table 4, by setting the uncertainty threshold to meet the 90% accuracy requirement, our model with data augmentation yielded a high accuracy of 92.8% against manual labels. By adjusting the threshold to target 85% accuracy, the achieved accuracy was 86.2%. Although these numbers are a bit lower than those without augmentation, the proportion of confident predictions is substantially increased (e.g., from 20% to 60.5% for the 90% accuracy requirement). This increase expands the potential for clinical applications, where more confidence predictions would be employed to assist clinicians in decision-making.

Table 4 Accuracy Relative to Manual Labels for Confident Predictions

|  | No Augmentation | | | | Data Augmentation | | | |
|---|---|---|---|---|---|---|---|---|
| Accuracy Requirement | 90% | 85% | 80% | -(all test data) | **90%** | **85%** | **80%** | -(all test data) |
| Accuracy based on Manual Label | 98.4% | 96.6% | 89.0% | 55.4% | **92.8%** | **86.2%** | **77.3%** | 73.3% |
| Case proportion | 20.0% | 37.0% | 49.4% | 100% | **60.5%** | **74.4%** | **90.4%** | 100% |

Table 5 summarizes the misclassified cases ('bad cases') among the model's confident predictions based on surrogated labels. Although there are 46 bad cases at the 90% accuracy threshold, all with surrogate label '1', only one prediction was actually incorrect according to manual labels. Interestingly, this single case received consistent judgment from one of our clinicians. Lowering the threshold to 80% produced similarly robust results: of the 166 cases deemed confident, 144 were accurate according to manual labels. These findings show that while surrogate labels may not perfectly align with manual annotations, our model reliably identifies auto-contour quality in low-uncertainty conditions, thereby enhancing the reliability of its outputs. Examples for correct predictions and bad cases of the auto-contours relative to GT contours are shown in Appendix.

Table 5 Bad Cases Among Confident Predictions

| Uncertainty Level | Surrogate Label | Model Predicted label | Manual label | # of Cases |
|---|---|---|---|---|
| Uncertainty Threshold for 90% Accuracy | 1 | **2** | **2** | **45** |
|  |  | 0 | 1(2 votes for 1;1 vote for 0) | 1 |
|  | 0 | 1 | 2 | 8 |
|  | 1 | 0 | 1 | 2 |

| | | 2 | 2 | **144** |
| Uncertainty Threshold for 80% Accuracy | | 2 | 1 | 4 |
| | 2 | 1 | 2 | 8 |

### B. *Model Developed with Manual Labels*

#### 1) *Transfer Learning with Limited Manual Labels*

Table 6 illustrates the influence of training sample size on the overall accuracy of transfer learning from our BOC model pre-trained using surrogate labels. The results indicate an initially positive trend, with accuracy increasing as the sample size grows from 10 to 30. When the training sample size reached 30 subjects (1252 slices), the model achieved an overall accuracy of 92.9%. Further increasing the sample size from 30 to 80 provided no performance gain. As shown in panels a1 and a3 of Fig. 5, the prediction accuracy decreased with increasing uncertainty. By applying our calibration strategy with an uncertainty threshold of $\tau = 0.33$ to meet the 90% accuracy requirement, the model attained 93.6% accuracy for 98.2% of all test cases with low uncertainty.

When transferred learning was employed using as few as 20 or more subjects, our model consistently achieved over 90% accuracy for more than 95% of the test cases (ranging from 96.5% to 98.7%) after applying the designated uncertainty threshold. These findings demonstrate that transferring from our pre-trained BOC model using limited manual labels can substantially improve performance. From the application perspective, over 95% of the testing cases exhibited confident predictions with accuracy exceeding 90%.

Table 6 Model development with manual labels

| # of Subjects for training (calibration) | Transfer Learning | | | | Train from Scratch | | | |
|---|---|---|---|---|---|---|---|---|
| | Overall | 90% Accuracy Requirement | | | Overall | 90% Accuracy Requirement | | |
| | Acc.[a] | Uncertainty Threshold | Acc.[b] | Case Proportion | Acc.[a] | Uncertainty Threshold | Acc.[b] | Case Proportion |
| 10(10) | 88.1% | 0.28 | 94.6% | 81.8 % | - | - | - | - |
| 20(20) | 91.6% | 0.30 | 93.0% | 96.5% | - | - | - | - |
| 30(34) | 92.9% | 0.33 | 93.6% | 98.2% | - | - | - | - |
| 40(34) | 91.4% | 0.35 | 92.3% | 98.7% | 81.1% | 0.15 | 93.0% | 55.2% |
| 50(34) | 92.2% | 0.33 | 93.2% | 98.1% | 83.3% | 0.17 | 91.3% | 67.4% |
| 80(34) | 92.1% | 0.30 | 93.3% | 97.1% | 85.6% | 0.19 | 89.9% | 83.1% |

a. Overall Acc.: Accuracy for all test data
b. Acc.: Accuracy for confident predictions (uncertainty below threshold) of test data

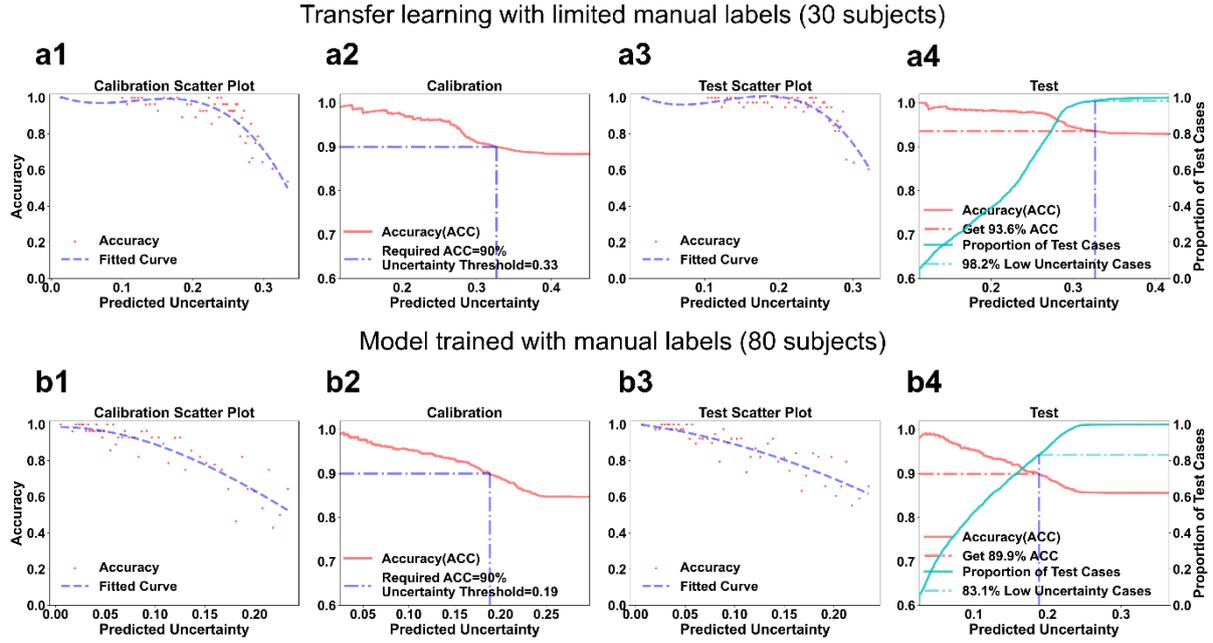

**Fig. 5 Accuracy vs. Uncertainty on Calibration and Test Data of Model Trained with Manual Labels.** a1-a4) Transfer learning from pre-trained BOC model with limited manually labelled data (30 subjects), from model trained with surrogate label and data augmentation. a1 and a3 show accuracy decreases with increasing uncertainty for both calibration and test data sets. a2 illustrates how to get the uncertainty threshold based on the 'accuracy vs. uncertainty' curve. a4 exhibits utilizing the uncertainty threshold 0.33 from calibration, we get 93.6% accuracy for 98.2% cases with uncertainty below the threshold. b1-b4) Train the BOC model with manually labelled data only (80 subjects), each subplot shows similar information but different uncertainty and accuracy.

### 2) Training with Manual Labels from Scratch

Increasing the sample size clearly improves our model's performance when trained from scratch using manual labels. Specifically, expanding the training sample size to 80 subjects increased the model's accuracy to 85.6%, though it remained lower than the accuracy achieved with transfer learning (see Table 6). Even so, employing our uncertainty-guided decision-making strategy continues to enhance the reliability of predictions. As demonstrated in Fig. 5 (panels b1-b4), focusing on confident predictions where the uncertainty is below the threshold set for 90% accuracy requirement yielded an accuracy of 89.9% for 83.1% of the test cases.

In conclusion, training our BOC model with a limited number of manual labels via transfer learning not only outperforms previous models trained using surrogate labels, but also increases the proportion of confident predictions. Notably, as few as 20–30 manual labeled subjects (each subject has multiple 2D slices) are sufficient to achieve satisfactory performance.

### C. Application

To assist clinicians in accelerating the contouring process for RT, Adaptive Radiotherapy (ART), and OART treatment planning, we developed BOC models. These models expedite decision-making by presenting confident predictions based on an uncertainty threshold predefined by clinical accuracy requirement. Predictions are then compared to clinicians' initial assessments. When the model's prediction aligns with the clinician's assessment, no warning is issued. However, if the model predicts 'Major Revision' or 'Not Acceptable' for a case that the clinician deems acceptable, a warning is triggered to prompt further review and mitigate potential errors.

In clinical applications, accurately identifying cases that require no revision enhances efficiency, whereas accurately flagging those that need major revision, or rejection ensures reliability in treatment planning. Accordingly, the QA process should prioritize accurate identification of Class 2 cases, that is, those deemed acceptable without revision (Class 2) versus those requiring major revision or rejection (Class 1 or Class 0). Table 7 illustrates the performance of our model's confident predictions across different training strategies.

**Without Manually Labeled Data:** When trained on surrogate labels alone with data augmentatio, the BOC model refers 60.5% of auto-contour quality assessments to clinicians, achieving a weighted F1-score of 0.949 for Class 2.

**With Limited Data (64 Subjects):** Employing transfer learning with manually labeled auto-contours from 64 subjects (30 for training/validation and 34 for calibration) allows referral of 98.2% of auto-contour quality assessments to clinicians, achieving a weighted F1-score of 0.938 for Class 2.

**With Expanded Data (114 Subjects):** Training a model from scratch using manually labeled auto-contours from 114 subjects (80 for training/validation and 34 for calibration) enables referral of 83.1% of auto-contour quality assessments to clinicians, achieving a weighted F1-score of 0.912 for Class 2.

Finally, we note that training the BOC model with manual labels, either via transfer learning or from scratch, significantly increases the proportion of confident predictions for Class 2, albeit at a slight cost to the weighted F1-score, compared to the model trained with surrogate labels only. However, this reduction in performance is minimal when transfer learning is employed, underscoring its effectiveness for improving contour quality assessments in clinical workflows.

Table 7 Results for Confident Predictions for Application (Class 2)

|  | Surrogate-Label Model | Transfer learning Model | Train from scratch Model |
| --- | --- | --- | --- |
| Case Proportion | 60.5% | 98.2% | 83.1% |
| Weighted Precision | 0.953 | 0.941 | 0.931 |
| Weighted Recall | 0.949 | 0.936 | 0.900 |
| Weighted F1 | 0.949 | 0.938 | 0.912 |

## IV. DISCUSSION AND CONCLUSIONS

AI-based auto-segmentation for medical images has been extensively studied, but it still requires substantial effort to review and revise auto-contours to ensure clinical reliability. AI-assisted auto-contour quality assessment (QA) methods represent a crucial step toward integrating AI-based auto-segmentation models into clinical workflows. However, several challenges arise when implementing such methods: the absence of ground truth (GT) contours for real-time evaluation, limited availability of human-labeled data, and inherent uncertainty in AI model predictions. An ideal AI-based auto-contour QA solution for clinical use should operate independently of GT contours while consistently providing reliable predictions.

In this study, we introduced an auto-contour QA method which does not require GT contours for predictions, relying solely on CT images and auto-contours as inputs. This method supports multiple development scenarios. First, it can be trained using surrogate labels derived from geometric characteristics, alleviating the need for extensive manual labeling. Secondly, data augmentation techniques including affine transformations, spacing alterations, rotations, and scaling further enhance the model performance. Last, our model can incorporate transfer learning to effectively utilize a small set of manually labeled data, substantially improving prediction performance.

To enhance prediction reliability, we developed a predictive uncertainty quantification method for our BOC model. Our results indicate a strong positive correlation between the model's estimated uncertainty and the uncertainty derived from clinicians' labels. Furthermore, cases identified by the BOC model as high-uncertainty were likewise challenging for clinicians, underscoring the effectiveness of our uncertainty estimation. By incorporating an additional calibration step, our method can achieve clinical accuracy exceeding 90% as required for confident predictions.

Clinically, our BOC models expedite decision-making by providing confident predictions that assist clinicians in their decision-making process, streamlining it to be faster and more accurate. When the model's prediction aligns with clinicians' initial assessment, no warnings is issued. However, if the model flags a need for major revisions or a rejection that the clinician deems acceptable, an alert prompts further review to swiftly address possible oversights. As demonstrated under various training strategies, the models' performance depends significantly on the effort invested in data labeling and the desired clinical accuracy level. Balancing these factors is crucial for optimizing clinical utility of our BOC models in OART planning. The models' robust ability in identifying 'No Revision' cases across diverse scenarios highlights their potential for practical application. In our framework, although the BOC models do not directly present contour quality in low-confidence cases, they can still provide valuable clinical support by flagging these cases. For example, a brief message in the user interface, such as "This case exhibits high uncertainty for the AI model; please review carefully", can alert clinicians to review the case with additional caution.

Despite these promising results, our study has limitations. First, AI-assisted auto-contour QA models need to be organ-specific, requiring separate training for different OARs or target structures, each with potentially different clinical requirements. Moreover, if the BOC model is trained using surrogate labels, the geometric thresholds for surrogate label generation must be tailored accordingly. This work focuses on rectum auto-contour QA primarily to demonstrate feasibility. Second, the efficiency gained and clinical time saved by our method depend highly on the performance of both the auto-segmentation and BOC models. High-performing auto-segmentation models typically generate more 'No Revision' cases, thus reducing the need for extensive manual revision. Simultaneously, more accurate BOC models facilitate quicker clinician decision-making by identifying a larger proportion of confident cases. Finally, a more effective auto-segmentation model may also produce a larger proportion of Class 2 ('No Revision') auto-contours, potentially causing class imbalance in downstream BOC model training. Additional strategies may be needed to address these challenges.

# APPENDIX

**Auto-segmentation model performance**

Three experienced clinicians reviewed the auto-contours for the test cases (1922 slices). As shown in Table A1, based on the majority voting results, 93.1% of the auto-contours were classified as Class 2 ('No/Minor revision'), 4.4% as Class 1 ('Major Revision'), and 2.5% as Class 0 ('Not Acceptable'). These results indicate that our in-house trained auto-segmentation model performs well, with over 90% of the auto-contours being suitable for clinical use. This high level of performance suggests that the model, if deployed in a clinical setting, could significantly assist clinicians. The examples of our model's prediction vs. clinicians' manual labels are shown in Fig A1.

The auto-contour quality distributions from the three clinicians are similar but we also observed inter-observer variations in the evaluation of auto-contour quality, as shown in Table A1. These variations highlight the inherent data uncertainty in auto-contour quality assessments. Additionally, we investigated the correlation between manual review-based contour quality, determined by majority voting, and geometric metrics. The Pearson correlation coefficients between manual evaluations and geometric metrics were 0.71 for DSC, -0.69 for HD95, and 0.45 for SDSC.

Table A1 Manual Reviewed Auto-contour Quality Distribution for Test Cases

| Class | Majority Voting Label | Clinician 1's Label | Clinician 2's Label | Clinician 3's Label |
|---|---|---|---|---|
| 0 | 49(2.5%) | 46(2.4%) | 66(3.4%) | 50(2.6%) |
| 1 | 84(4.4%) | 133(6.9%) | 39(2.0%) | 109(5.7%) |
| 2 | 1789(93.1%) | 1743(90.7%) | 1817(94.5%) | 1763(91.7%)) |

Fig A1 shows examples for correct and incorrect auto-contour quality predictions among the confident cases. Taking panel e as an example, there is a significant discrepancy between the auto-contour (red) and the ground truth contour (blue) in the posterior region, prompting the model to confidently classify it as Class 0 (`Rejection'). However, clinicians assessed it as usable (Class 1, `Major Revision').

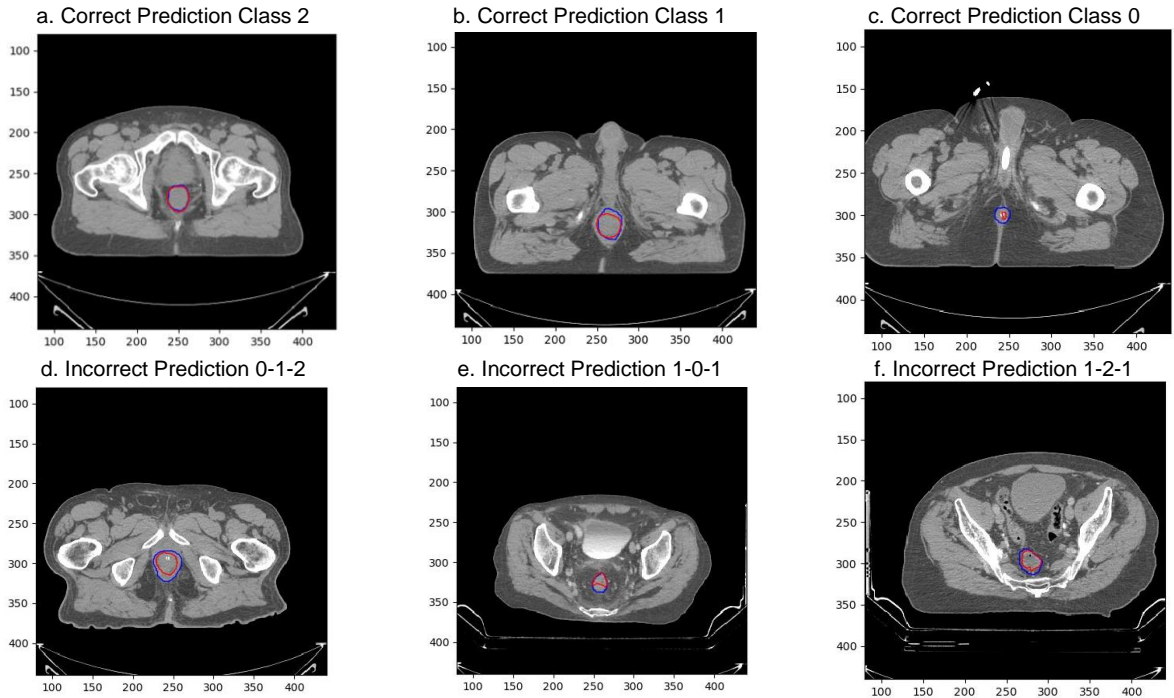

**Fig A1 Predicted Results vs. Manual Labels.** Blue Contour: Ground Truth contour; Red Contour: Auto-contour. a-c: Correct contour quality predictions for different labels in confident predictions. d-f: Wrong predictions for different labels in confident predictions; '1-0-1' means classes for surrogate label "1"-model predicted label "0"-manual label "1", similarly for other incorrect predictions.

# REFERENCE

[1] A. M. Grilo, B. Santos, I. Baptista, and F. Monsanto, "Exploring the cancer patients' experiences during external radiotherapy: A systematic review and thematic synthesis of qualitative and quantitative evidence," *Eur. J. Oncol. Nurs.*, vol. 52, p. 101965, Jun. 2021, doi: 10.1016/j.ejon.2021.101965.

[2] R. L. Christiansen *et al.*, "Online adaptive radiotherapy potentially reduces toxicity for high-risk prostate cancer treatment," *Radiother. Oncol.*, vol. 167, pp. 165–171, Feb. 2022, doi: 10.1016/j.radonc.2021.12.013.

[3] J. Duan *et al.*, "Evaluating the clinical acceptability of deep learning contours of prostate and organs-at-risk in an automated prostate treatment planning process," *Med. Phys.*, vol. 49, no. 4, pp. 2570–2581, Apr. 2022, doi: 10.1002/mp.15525.

[4] M. V. Sherer *et al.*, "Metrics to evaluate the performance of auto-segmentation for radiation treatment planning: A critical review," *Radiother. Oncol.*, vol. 160, pp. 185–191, Jul. 2021, doi: 10.1016/j.radonc.2021.05.003.

[5] M. B. Altman *et al.*, "A framework for automated contour quality assurance in radiation therapy including adaptive techniques," *Phys. Med. Biol.*, vol. 60, no. 13, pp. 5199–5209, Jul. 2015, doi: 10.1088/0031-9155/60/13/5199.

[6] J. Duan *et al.*, "Contouring quality assurance methodology based on multiple geometric features against deep learning auto-segmentation," *Med. Phys.*, vol. 50, no. 5, pp. 2715–2732, May 2023, doi: 10.1002/mp.16299.

[7] Y. Zhang *et al.*, "Comprehensive Clinical Usability-Oriented Contour Quality Evaluation for Deep Learning Auto-segmentation: Combining Multiple Quantitative Metrics Through Machine Learning," *Pract. Radiat. Oncol.*, vol. 15, no. 1, pp. 93–102, Jan. 2025, doi: 10.1016/j.prro.2024.07.007.

[8] K. Men, H. Geng, T. Biswas, Z. Liao, and Y. Xiao, "Automated Quality Assurance of OAR Contouring for Lung Cancer Based on Segmentation With Deep Active Learning," *Front. Oncol.*, vol. 10, p. 986, Jul. 2020, doi: 10.3389/fonc.2020.00986.

[9] X. Chen *et al.*, "CNN-Based Quality Assurance for Automatic Segmentation of Breast Cancer in Radiotherapy," *Front. Oncol.*, vol. 10, p. 524, Apr. 2020, doi: 10.3389/fonc.2020.00524.

[10] C. A. T. Van Den Berg and E. F. Meliadò, "Uncertainty Assessment for Deep Learning Radiotherapy Applications," *Semin. Radiat. Oncol.*, vol. 32, no. 4, pp. 304–318, Oct. 2022, doi: 10.1016/j.semradonc.2022.06.001.

[11] K. Wang, M. Dohopolski, Q. Zhang, D. Sher, and J. Wang, "Towards reliable head and neck cancers locoregional recurrence prediction using delta-radiomics and learning with rejection option," *Med. Phys.*, vol. 50, no. 4, pp. 2212–2223, 2023, doi: 10.1002/mp.16132.

[12] M. Dohopolski, L. Chen, D. Sher, and J. Wang, "Predicting lymph node metastasis in patients with oropharyngeal cancer by using a convolutional neural network with associated epistemic and aleatoric uncertainty," *Phys. Med. Biol.*, vol. 65, no. 22, p. 225002, Nov. 2020, doi: 10.1088/1361-6560/abb71c.

[13] M. Dohopolski *et al.*, "Uncertainty estimations methods for a deep learning model to aid in clinical decision-making – a clinician's perspective".

[14] A. Balagopal *et al.*, "A deep learning-based framework for segmenting invisible clinical target volumes with estimated uncertainties for post-operative prostate cancer radiotherapy," *Med. Image Anal.*, vol. 72, p. 102101, Aug. 2021, doi: 10.1016/j.media.2021.102101.

[15] Y. Kwon, J.-H. Won, B. J. Kim, and M. C. Paik, "Uncertainty quantification using Bayesian neural networks in classification: Application to biomedical image segmentation," *Comput. Stat. Data Anal.*, vol. 142, p. 106816, Feb. 2020, doi: 10.1016/j.csda.2019.106816.

[16] A. Jungo, R. Meier, E. Ermis, E. Herrmann, and M. Reyes, "Uncertainty-driven Sanity Check: Application to Postoperative Brain Tumor Cavity Segmentation," Jun. 08, 2018, *arXiv*: arXiv:1806.03106. Accessed: Apr. 17, 2024. [Online]. Available: http://arxiv.org/abs/1806.03106

[17] "Using Spatial Probability Maps to Highlight Potential Inaccuracies in Deep Learning-Based Contours: Facilitating Online Adaptive Radiation Therapy - ClinicalKey." Accessed: Apr. 17, 2024. [Online]. Available: https://www.clinicalkey.com/#!/content/playContent/1-s2.0-S2452109421000166?returnurl=https:%2F%2Flinkinghub.elsevier.com%2Fretrieve%2Fpii%2FS2452109421000166%3Fshowall%3Dtrue&referrer=

[18] D. Nguyen *et al.*, "A comparison of Monte Carlo dropout and bootstrap aggregation on the performance and uncertainty estimation in radiation therapy dose prediction with deep learning neural networks," *Phys. Med. Biol.*, vol. 66, no. 5, p. 054002, Feb. 2021, doi: 10.1088/1361-6560/abe04f.

[19] F. J. S. Bragman *et al.*, "Uncertainty in Multitask Learning: Joint Representations for Probabilistic MR-only Radiotherapy Planning," in *Medical Image Computing and Computer Assisted Intervention – MICCAI 2018*, A. F. Frangi, J. A. Schnabel, C. Davatzikos, C. Alberola-López, and G. Fichtinger, Eds., Cham: Springer International Publishing, 2018, pp. 3–11. doi: 10.1007/978-3-030-00937-3_1.

[20] M. F. Spadea, M. Maspero, P. Zaffino, and J. Seco, "Deep learning based synthetic-CT generation in radiotherapy and PET: A review," *Med. Phys.*, vol. 48, no. 11, pp. 6537–6566, 2021, doi: 10.1002/mp.15150.

[21] M. Hemsley *et al.*, "Deep Generative Model for Synthetic-CT Generation with Uncertainty Predictions," in *Medical Image Computing and Computer Assisted Intervention – MICCAI 2020*, A. L. Martel, P. Abolmaesumi, D. Stoyanov, D. Mateus, M. A. Zuluaga, S. K. Zhou, D. Racoceanu, and L. Joskowicz, Eds., Cham: Springer International Publishing, 2020, pp. 834–844. doi: 10.1007/978-3-030-59710-8_81.

[22] Y. Gal, "Uncertainty in Deep Learning," University of Cambridge. [Online]. Available: chrome-extension://efaidnbmnnnibpcajpcglclefindmkaj/https://www.cs.ox.ac.uk/people/yarin.gal/website/thesis/thesis.pdf

[23] X. Shi, W. Cao, and S. Raschka, "Deep neural networks for rank-consistent ordinal regression based on conditional probabilities," *Pattern Anal. Appl.*, vol. 26, no. 3, pp. 941–955, Aug. 2023, doi: 10.1007/s10044-023-01181-9.

[24] N. Bakx *et al.*, "Clinical evaluation of a deep learning segmentation model including manual adjustments afterwards for locally advanced breast cancer," *Tech. Innov. Patient Support Radiat. Oncol.*, vol. 26, p. 100211, May 2023, doi: 10.1016/j.tipsro.2023.100211.

[25] H. Baroudi *et al.*, "Automated Contouring and Planning in Radiation Therapy: What Is 'Clinically Acceptable'?," *Diagnostics*, vol. 13, no. 4, p. 667, Feb. 2023, doi: 10.3390/diagnostics13040667.

[26] J. Duan *et al.*, "Incremental retraining, clinical implementation, and acceptance rate of deep learning auto-segmentation for male pelvis in a multiuser environment," *Med. Phys.*, vol. 50, no. 7, pp. 4079–4091, Jul. 2023, doi: 10.1002/mp.16537.

[27] J. Duan *et al.*, "Evaluating the clinical acceptability of deep learning contours of prostate and organs-at-risk in an automated prostate treatment planning process," *Med. Phys.*, vol. 49, no. 4, pp. 2570–2581, 2022, doi: 10.1002/mp.15525.

[28] E. Cha *et al.*, "Clinical implementation of deep learning contour autosegmentation for prostate radiotherapy," *Radiother. Oncol.*, vol. 159, pp. 1–7, Jun. 2021, doi: 10.1016/j.radonc.2021.02.040.

[29] S. Nikolov *et al.*, "Deep learning to achieve clinically applicable segmentation of head and neck anatomy for radiotherapy," Jan. 13, 2021, *arXiv*: arXiv:1809.04430. Accessed: Apr. 11, 2024. [Online]. Available: http://arxiv.org/abs/1809.04430

[30] T. Islam, Md. S. Hafiz, J. R. Jim, Md. M. Kabir, and M. F. Mridha, "A systematic review of deep learning data augmentation in medical imaging: Recent advances and future research directions," *Healthc. Anal.*, vol. 5, p. 100340, Jun. 2024, doi: 10.1016/j.health.2024.100340.

[31] Y. Gal and Z. Ghahramani, "Dropout as a Bayesian Approximation: Representing Model Uncertainty in Deep Learning," in *Proceedings of The 33rd International Conference on Machine Learning*, PMLR, Jun. 2016, pp. 1050–1059. Accessed: Feb. 23, 2025. [Online]. Available: https://proceedings.mlr.press/v48/gal16.html